\documentclass[runningheads]{llncs}
\usepackage[T1]{fontenc}
\usepackage{booktabs}
\usepackage{graphicx}
\usepackage{arydshln}
\usepackage{array}
\usepackage{pifont}
\usepackage{xcolor}

\usepackage{float}
\usepackage{placeins}
\newcommand{\cmark}{\textcolor{green!80!black}{\ding{51}}}
\newcommand{\xmark}{\textcolor{red}{\ding{55}}}

\begin{document}
\title{From Videos to Conversations: Egocentric Instructions for Task Assistance}
%
%

\author{Lavisha Aggarwal\inst{1} \and Vikas Bahirwani\inst{1} \and Andrea Colaco\inst{1}}

\authorrunning{L. Aggarwal et al.}
%
\institute{Google, USA}
\maketitle              
\begin{abstract}
Many everyday tasks, ranging from appliance repair and cooking to car maintenance, require expert knowledge, particularly for complex, multi-step procedures. Despite growing interest in AI agents for augmented reality (AR) assistance, progress remains limited by the scarcity of large-scale multimodal conversational datasets grounded in real-world task execution, in part due to the cost and logistical complexity of human-assisted data collection. In this paper, we present a framework to automatically transform single-person instructional videos into
two-person multimodal task-guidance conversations. Our fully automatic pipeline, based on large language models, provides a scalable and cost-efficient alternative to traditional data collection approaches. Using this framework, we introduce HowToDIV, a multimodal dataset comprising 507 conversations, 6,636 question-answer pairs, and 24 hours of video spanning multiple domains. Each session consists of a multi-turn expert-novice interaction. Finally, we report baseline results using Gemma-3 and Qwen 2.5 on HowToDIV, providing an initial benchmark for multimodal procedural task assistance.

\end{abstract}
\section{Introduction}
\label{sec:intro}

Recent advancements in general-purpose AI agents have enabled new forms of human assistance across a wide range of everyday activities. A prominent class of such activities involves procedural tasks\footnote{A procedural task comprises a structured sequence of steps aimed at achieving a specified goal, allowing limited permissible orderings while treating other deviations as execution errors.}, such as furniture assembly, cooking, mechanical repair, which require executing a sequence of interdependent steps to achieve a well-defined goal. These tasks are difficult to memorize and execute reliably due to their length, fine-grained details, and sensitivity to execution order and errors.
\setlength{\textfloatsep}{8pt}
\setlength{\intextsep}{8pt}
\begin{figure}[ht]
    \centering
    \includegraphics[width=1\textwidth]{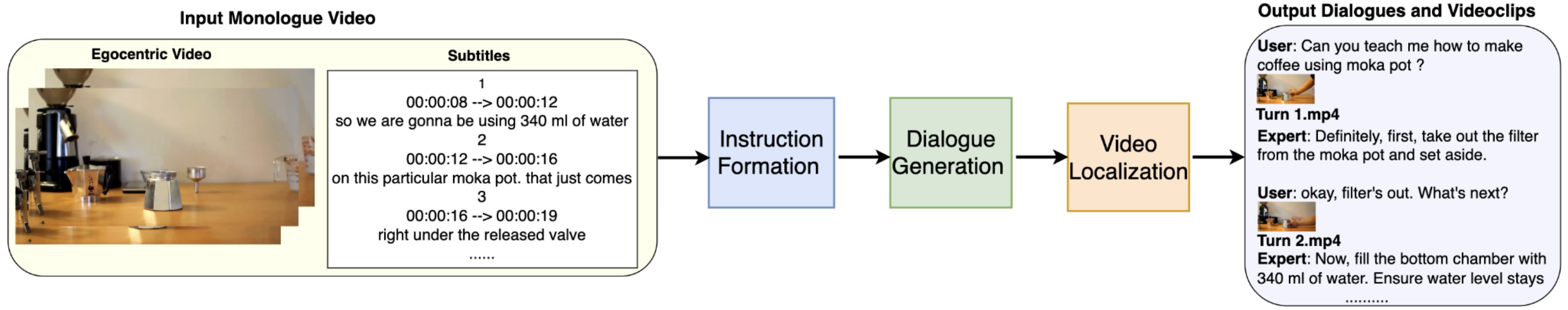}
    \caption{
    Overview of the proposed MDC (Monologue-to-Dialogue Conversion)  pipeline. An instructional monologue video with expert subtitles is processed to infer procedural instructions, generate multi-turn expert-user dialogues using LLMs, and temporally align video segments with user turns.}
    \label{fig:overview}
\end{figure}
The growing availability of augmented reality (AR) wearables presents a compelling opportunity to develop vision-language agents capable of providing real-time, step-by-step procedural guidance. In such settings, an AI assistant can observe the user’s actions, provide instructions incrementally, and proactively detect and correct execution errors. However, progress in this direction is constrained by the lack of multi-modal conversational datasets grounded in real-world task execution, which limits both model training and agentic evaluation.

Datasets capturing expert-novice interactions in procedural tasks are particularly valuable for this problem setting. In a typical AR task-assistance scenario, a novice user equipped with a wearable device streams audio-visual input to a vision-language model \cite{gemma3}, which must reason over procedural knowledge, track task progress, answer intermediate questions, and adapt guidance based on observed user behavior. Collecting such datasets through human-facilitated interactions, however, is expensive, time-consuming, and difficult to scale, as it requires in-field recording, supervision, and annotation across diverse tasks.

Existing datasets only partially address these requirements. For example, HoloAssist \cite{holoassist} provides egocentric video and dialogue for collaborative object manipulation tasks captured using HoloLens. While valuable, it is limited in tasks and required hundreds of hours of human effort, making it costly and restrictive in scope. More broadly, large-scale instructional video datasets—such as NIV \cite{niv}, EgoPER \cite{egoper}, YouCookII \cite{youcook2}, Assembly101 \cite{assembly101}, CrossTask \cite{crosstask}, EPIC-KITCHENS \cite{epickitchen}, and COIN \cite{coin}—are readily available but are not conversational. These datasets typically consist of monologue-style demonstrations intended for offline consumption, rather than interactive, real-time task assistance, and lack collaborative dialogue grounded in user actions.

Motivated by the high cost of human-assisted data collection and the abundance of instructional video corpora, we explore an automatic alternative for constructing procedural task-assistance datasets. We propose a prompting-based framework that leverages large language models to transform single-person instructional videos into two-person expert-novice dialogues with turn-level video grounding, enabling scalable dataset construction without manual intervention.

Using this framework, we introduce HowToDIV, a multimodal dataset built on top of the NIV \cite{niv} and EgoPER \cite{egoper} datasets and extensible to other instructional video sources. HowToDIV comprises multi-turn Dialogues paired with egocentric Video clips and step-level procedural Instructions. Although HowToDIV is generated automatically using LLMs, dialogue synthesis is constrained by real instructional videos, subtitles, an ordered set of atomic procedural steps, and turn-level egocentric grounding. For datasets such as EgoPER, user errors and expert corrections are explicitly conditioned on human-annotated error labels, preventing hallucinated actions or instructions.


We benchmark HowToDIV using the publicly available vision-language models \cite{gemma3}, \cite{qwen2.5}, evaluating performance with BLEU, ROUGE, and an LLM-as-a-Judge metric to establish baseline results for procedural task assistance. 


Unlike prior instructional video datasets or synthetic dialogue corpora, HowToDIV is the first dataset that jointly enforces (i) atomic procedural steps, (ii) multi-turn instructional dialogue, (iii) turn-level egocentric video grounding, and (iv) explicit modeling of user errors and expert corrections, all within a fully automatic and scalable pipeline. Our contributions\footnote{Our dataset and code are publicly available at
https://github.com/google/howtodiv.} are summarized as follows:
\begin{itemize}
 \item We propose a fully automatic instructional monologue-to-dialogue framework that integrates domain knowledge from expert-help videos with the linguistic reasoning capabilities of vision language models to generate structured conversations with explicit modeling of user errors and corrections.
\item We introduce HowToDIV, a multimodal conversational dataset comprising 507 sessions, 6,636 dialogue turns, and 24 hours of aligned video clips across varied tasks; with variations in user speech styles, action accuracy, and step orderings.
\item We establish baseline performance using Gemma-3 and Qwen-2.5, providing the first benchmark for multimodal task assistance on HowToDIV.
\end{itemize}

\section{Related Works}
\label{sec:related_works}

\textbf{Instructional Video Datasets.}
Instructional videos provide a rich substrate for learning complex, multi-step procedures, spanning domains such as cooking, home maintenance, and mechanical repair. Large-scale efforts such as HowTo100M~\cite{howto100m} have substantially advanced this area by collecting 136 million clips from over 1.22 million narrated instructional videos, covering more than 23,000 distinct tasks. By leveraging automatically transcribed narrations as weak supervision, HowTo100M enables scalable learning of text-video representations, offering an alternative to costly manual annotation required by smaller curated datasets such as MSR-VTT~\cite{msrvtt}, YouCook2~\cite{youcook2}, and EPIC-KITCHENS~\cite{epickitchen}.

Charades~\cite{charades} is another large-scale dataset capturing everyday indoor activities performed by users in their own homes. While it contains diverse human actions, its focus is on action description rather than task execution and lacks any notion of interactive dialogue. 
Other instructional datasets such as COIN~\cite{coin} and CrossTask~\cite{crosstask} align videos with WikiHow articles~\cite{wikihow} for step-level understanding. Although these datasets have been instrumental for research in action recognition, temporal segmentation, and procedural reasoning, they primarily consist of monologue-style instructions and do not model the interactive, multi-turn exchanges required for real-world task assistance. 

\textbf{Egocentric Video Datasets for Procedural Tasks.}
Egocentric video captures activities from a first-person perspective, offering direct insight into user intent, object manipulation, and action execution. This viewpoint has motivated large-scale datasets such as Ego4D~\cite{ego4d}, which spans thousands of hours of daily activities, as well as task-focused datasets including EPIC-KITCHENS~\cite{epickitchen} for cooking and Assembly101~\cite{assembly101} for object assembly.

More recent datasets emphasize human interaction within egocentric procedural settings. HoloAssist~\cite{holoassist} introduces a two-person collaborative scenario in which a performer wearing an AR headset is verbally guided by a remote instructor. The dataset provides rich multimodal signals—including RGB, depth, audio, gaze, and hand pose—along with annotations for actions, mistakes, and interventions, underscoring the importance of grounded guidance. EgoPER~\cite{egoper} further advances this line of work by introducing a fine-grained taxonomy of procedural errors in egocentric cooking tasks, capturing omission, addition, modification, and correction behaviors. ALFRED~\cite{alfred} offers long-horizon compositional household tasks in interactive environments, focusing on sequential planning and execution. 
Despite their strengths, these egocentric datasets do not natively support structured expert–novice dialogue aligned with step-wise procedural execution, instead emphasizing monologue-style demonstrations, action understanding, or error detection in isolation. 
\section{HowToDIV Dataset}
\label{sec:dataset}

We introduce HowToDIV, a multimodal dataset designed for two-person procedural task assistance, comprising expert-user dialogues, step-level instructions, and egocentric video clips aligned at the dialogue-turn level. Section~\ref{sec:data_curation} describes the data sources and curation process, followed by dataset annotations and statistics in Section~\ref{sec:data_statistics} and the automatic data generation pipeline detailed in Section~\ref{sec:method}.

\subsection{Data Curation}
\label{sec:data_curation}
HowToDIV consists of goal-oriented, multi-turn conversations between a novice user and an expert guide performing multi-step procedural tasks. The dataset contains 507 sessions, 6,636 dialogue turns, and approximately 24 hours of egocentric video, capturing the user’s viewpoint via a wearable camera. It spans three domains—cooking, mechanical repair, and planting—covering nine tasks.

Each session is initiated by a user requesting guidance for a specific task. The expert responds with step-by-step instructions, while the user executes actions incrementally, requests subsequent steps, and asks clarification questions as needed. Conversations terminate upon task completion. Each user turn is paired with a temporally aligned video clip depicting the performed action, and a summarized procedural instruction sequence is provided for each session.

HowToDIV is constructed on top of two publicly available instructional video datasets that feature single-person demonstrations: Narrated Instruction Videos (NIV) \cite{niv} and the Egocentric Procedural Error (EgoPER) dataset \cite{egoper}.

\textbf{Narrated Instruction Videos (NIV)}. The NIV dataset \cite{niv} consists of 150 YouTube videos across five tasks, with 30 videos per task and an average duration of approximately 2 minutes (or 4,000 frames). Each video depicts an expert narrating and demonstrating a complete task, covering complex human-object interactions in both indoor and outdoor environments. NIV provides English transcripts obtained via automatic speech recognition and corrected by human annotators. For HowToDIV, we manually select videos with an egocentric or near-egocentric viewpoint to better approximate a user-worn camera perspective.

\begin{figure*}[ht]
     \centering   
         \centering
         \includegraphics[width=1.0\textwidth]{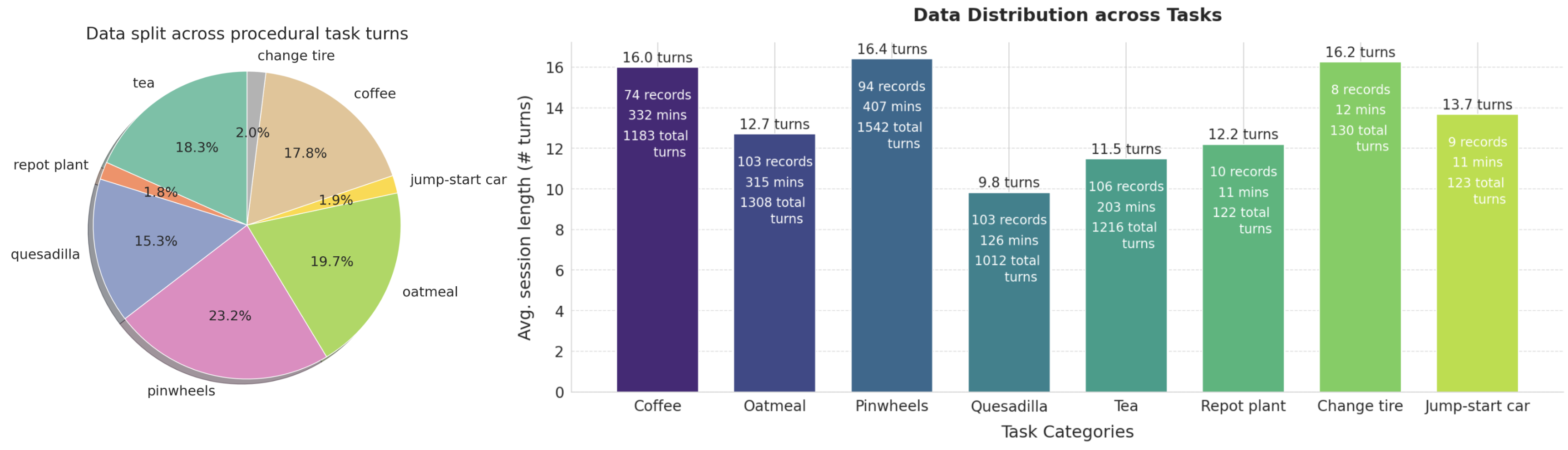}
         \caption{Distribution of HowToDIV across procedural activities. Left: Proportion of user-expert turns for various tasks. Right: Average conversation length as turn count (top of bars); the bars show conversation count, video duration, and QA pair count.}
         \label{fig:task_split}
\end{figure*}
\begin{table}[t]
\centering
\footnotesize
\setlength{\abovecaptionskip}{5pt}
\setlength{\tabcolsep}{4pt}
\fontsize{7.5}{8}\selectfont
\begin{minipage}[t]{0.48\linewidth}
\centering
\vspace{0pt}
\begin{tabular}{p{0.5\linewidth} p{0.5\linewidth}}
\toprule
\textbf{INSTRUCTION} & \textbf{USER ERROR} \\
\midrule
Place the tea bag in the mug & Dropped the tea bag on floor\\
Roll the tortilla into a wrap & Folded the tortilla in half \\
Fold the filter to create a semicircle & Tore the paper filter \\
\bottomrule
\end{tabular}
\end{minipage}
\hfill
\begin{minipage}[t]{0.48\linewidth}
\centering
\vspace{0pt}
\begin{tabular}{p{0.5\linewidth} p{0.5\linewidth}}
\toprule
\textbf{INSTRUCTION} & \textbf{MODIFICATION} \\
\midrule
Drizzle honey in bowl & Pour sugar instead of honey\\
Stir using spoon & Stir using knife \\
Slice tortilla using knife & Rip tortilla by hands \\
Place tortilla on cutting board & Place tortilla on table \\
\bottomrule
\end{tabular}
\end{minipage}
\setlength{\belowcaptionskip}{2pt}
\caption{Examples of (Left) User errors and (Right) Modifications and the corresponding instruction steps from HowToDIV.}
\label{tab:mistakes}
\end{table}

    

\textbf{Egocentric Procedural Error (EgoPER)}.
The EgoPER dataset \cite{egoper} comprises 386 egocentric procedural task videos collected using Microsoft HoloLens2 from 11 participants. It includes multimodal streams (RGB, depth, audio, gaze, and hand tracking) for five cooking tasks: pinwheels, quesadilla, oatmeal, coffee, and tea. Unlike NIV, EgoPER does not contain spoken narration; instead, it provides timestamped, fine-grained action annotations for each procedural step. The dataset includes both correct executions (213 videos) and erroneous executions (173 videos), totaling approximately 28 hours of footage. Errors are categorized into step omissions, additions, modifications, slips, and corrections. For HowToDIV, we retain all normal videos and selectively include erroneous videos involving step modifications and corrections, aligning with our goal of modeling expert intervention following user mistakes.


\setlength{\textfloatsep}{8pt}
\setlength{\intextsep}{4pt}
\begin{figure}[ht]
\centering
\begin{minipage}[t]{0.48\textwidth}
\centering
\includegraphics[width=\linewidth]{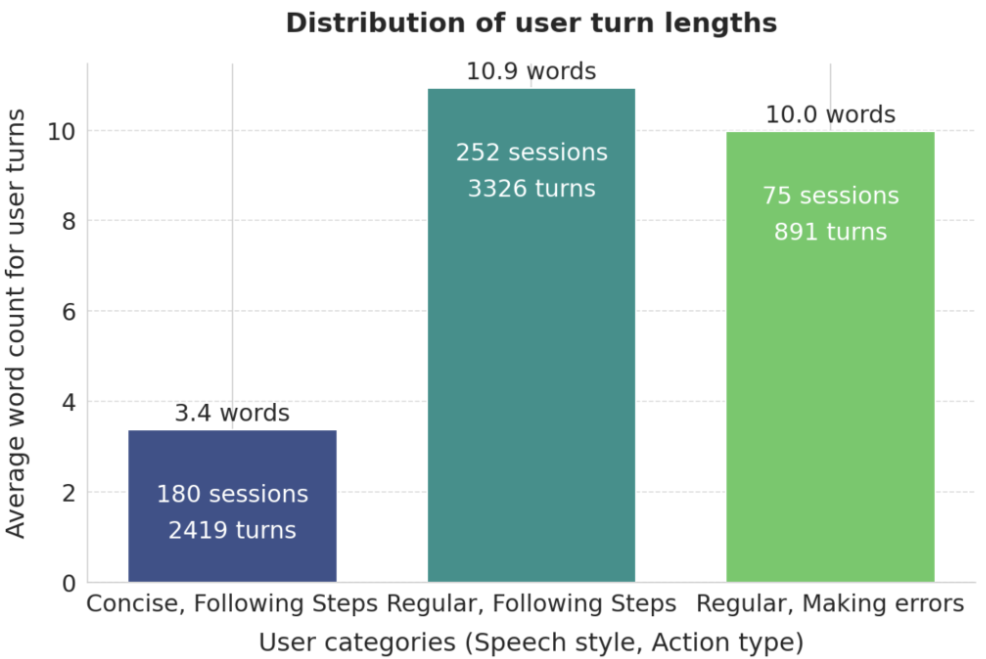}
\caption{Distribution of user dialogue lengths across different speech and activity categories. Category 1—Concise speech with correct actions (avg. 3.4 words); Category 2—Regular (or more verbose) speech with correct actions (avg. 10.9 words); Category 3—Regular speech with user errors (avg. 10 words).}
\label{fig:howtodiv_user_turnlen}
\end{minipage}
\hfill
\begin{minipage}[t]{0.48\textwidth}
\centering
\includegraphics[width=\linewidth]{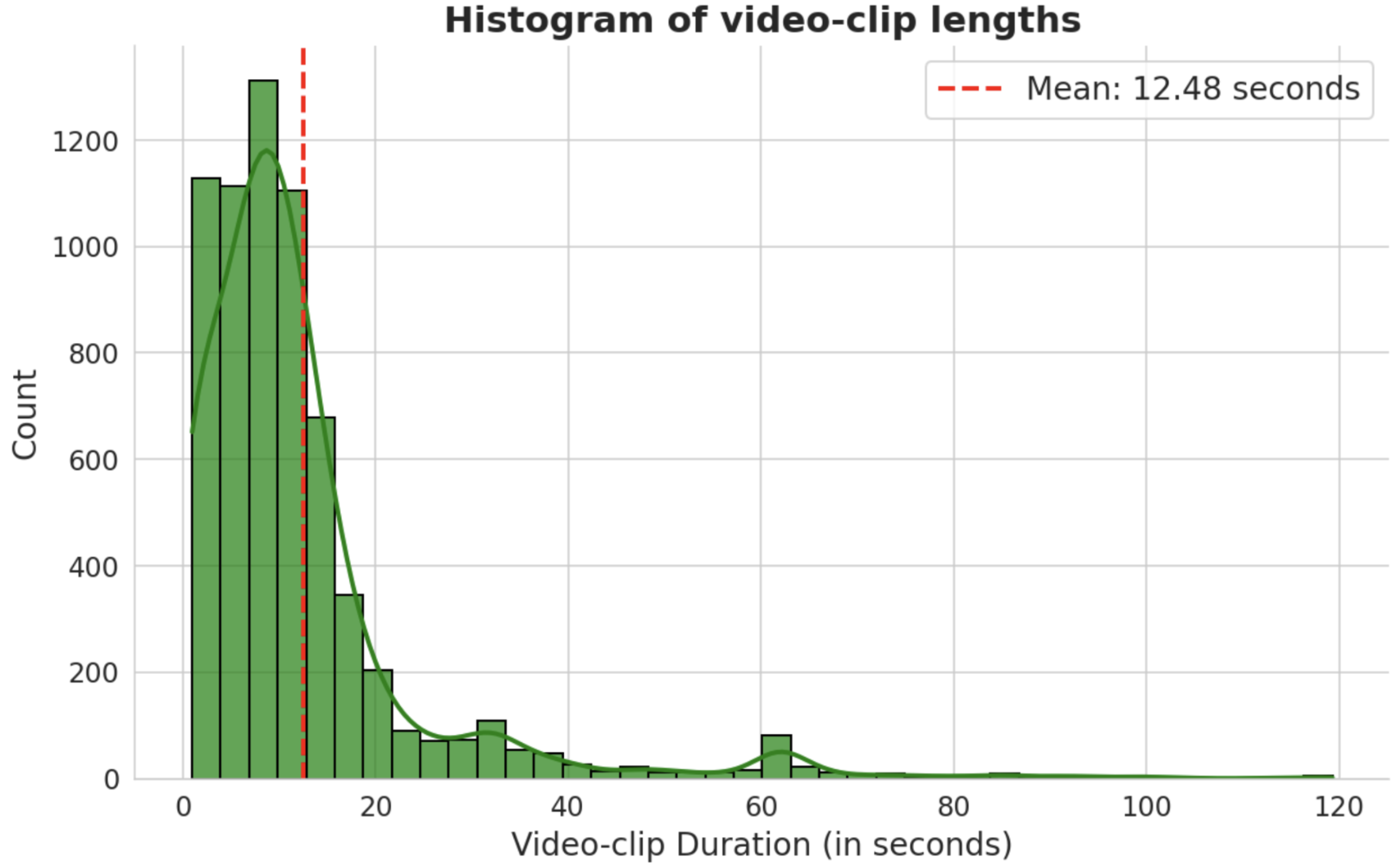}
\caption{Histogram of user-turn video lengths, with an average duration of 12.48 seconds.}
\label{fig:vid_len}
\end{minipage}

\end{figure}

\subsection{Annotations and Statistics}
\label{sec:data_statistics}
\begin{table*}[t]
    \setlength{\abovecaptionskip}{6pt}
    \centering
    \resizebox{.95\textwidth}{!}{
    \begin{tabular}{c|c|cccc|ccc}
    \toprule   
     
     Dataset & Domain & Dialogues & Temporal-steps & Narrations & Instructions & Errors & Egocentric & Videos \\
     \midrule
     NIV~\cite{niv} & Varied & \xmark & \xmark & \cmark & \xmark & \xmark & \xmark & 6.66\\
     HowTo100M~\cite{howto100m} & Varied & \xmark & \xmark & \cmark & \xmark & \xmark & \xmark & 134.4k\\
     HoloAssist~\cite{holoassist} & Assembly & \cmark & \cmark & \xmark & \cmark & \cmark & \cmark & 2221\\
     EpicTent~\cite{epictent} & Tent Making & \xmark & \cmark & \xmark & \cmark & \cmark & \cmark & 7\\
     EgoPER~\cite{egoper} & Cooking & \xmark & \cmark & \xmark & \cmark & \cmark & \cmark & 28\\
     EGTEA~\cite{egtea} & Cooking & \xmark & \cmark & \xmark & \xmark & \xmark & \cmark & 28\\
     Assembly101~\cite{assembly101} & Toy-assembly & \xmark & \cmark & \xmark & \cmark & \cmark & \cmark & 167\\
    
     COIN~\cite{coin} & Varied & \xmark & \cmark & \cmark & \xmark & \xmark & \xmark & 476\\
     
     CrossTask~\cite{crosstask} & Varied & \xmark & \cmark & \cmark & \cmark & \xmark & \xmark & 376\\
     YouCook2~\cite{youcook2} & Cooking & \xmark & \cmark & \cmark & \cmark & \xmark & \xmark & 176\\
      EPIC-KITCHENS~\cite{epickitchen} & Cooking & \xmark & \cmark & \cmark & \xmark & \xmark & \cmark & 100\\
     \midrule
     
     HowToDIV (Ours) & Varied & \cmark & \cmark & \cmark & \cmark & \cmark & \cmark & 576\\
    \bottomrule
    \end{tabular}
    }
\caption{Comparison of HowToDIV with existing instructional video and dialogue datasets for procedural tasks.}
\label{tab:dataset_comparison}
\end{table*}

The distribution of sessions across task categories is shown in Fig.~\ref{fig:task_split}. Average number of turns per session ranges from 9.8 turns for simpler tasks to 16.4 turns for more complex ones, with an overall mean of 13 turns.

HowToDIV is annotated along user speech-style and action-type dimensions (Fig.~\ref{fig:howtodiv_user_turnlen}). User dialogues are categorized into Concise and Regular speech styles. The dataset includes 180 concise sessions (average 3.4 words per user turn) and 327 regular sessions (average 10.7 words per user turn). Regular speech sessions are further divided into Following-Steps (252 sessions), where users execute instructions correctly, and Making-Errors (75 sessions), where users deviate from the prescribed procedure and are subsequently corrected by the expert. Representative error types and step modifications are summarized in Tab.~\ref{tab:mistakes}.

Expert responses average 19.3 words per turn. Since each user turn is paired with an egocentric video clip, clip durations vary widely, ranging from 1 second to over 2 minutes, with an average length of 12.5 seconds (Fig.~\ref{fig:vid_len}). A comparison with existing instructional and egocentric datasets is provided in Tab.~\ref{tab:dataset_comparison}. To the best of our knowledge, HowToDIV is the first dataset to jointly provide procedural instructions, expert-user dialogues, and egocentric video clips, constructed in a fully automatic and cost-efficient manner, making it suitable for scalable research in AR-based procedural task assistance.

\subsection{Dataset Quality Control}
Since HowToDIV is generated automatically, we perform both human and automatic quality checks. We randomly sample 175 dialogue turns and have two trained annotators evaluate instruction correctness, dialogue naturalness, and video-step alignment using a binary usable/not-usable label. A turn is considered usable if both annotators agree it satisfies at least two criteria, under which \textbf{93.2\%} of sampled turns are judged usable. Additionally, we apply automatic filtering for duplicates, degenerate responses, length outliers, profanity, and failed temporal localization, removing fewer than \textbf{4\%} of initially generated turns. While smaller than web-scale instructional corpora, HowToDIV offers substantially higher structural density per sample, with each dialogue grounded in egocentric video and procedural state.

\section{Method}
\label{sec:method}

We propose a prompt-engineering-based framework for transforming egocentric instructional videos into multi-turn conversations. We formalize this process as an algorithmic transformation that converts a monologue instructional video into a structured assistive dialogue with turn-level video grounding. We refer to this transformation as the \textit{Monologue-to-Dialogue Conversion (MDC)} pipeline. Our approach leverages a pre-trained multimodal large language model (LLM) with frozen weights and rests on two key observations: (i) egocentric instructional videos encode sufficient visual and linguistic cues to recover a complete, step-wise task specification, and (ii) a structured instruction set can be systematically converted into an interactive teaching dialogue that mirrors real-world procedural assistance.  
\begin{figure*}[ht]
     \centering 
         \centering
         \includegraphics[width=1.0\textwidth]{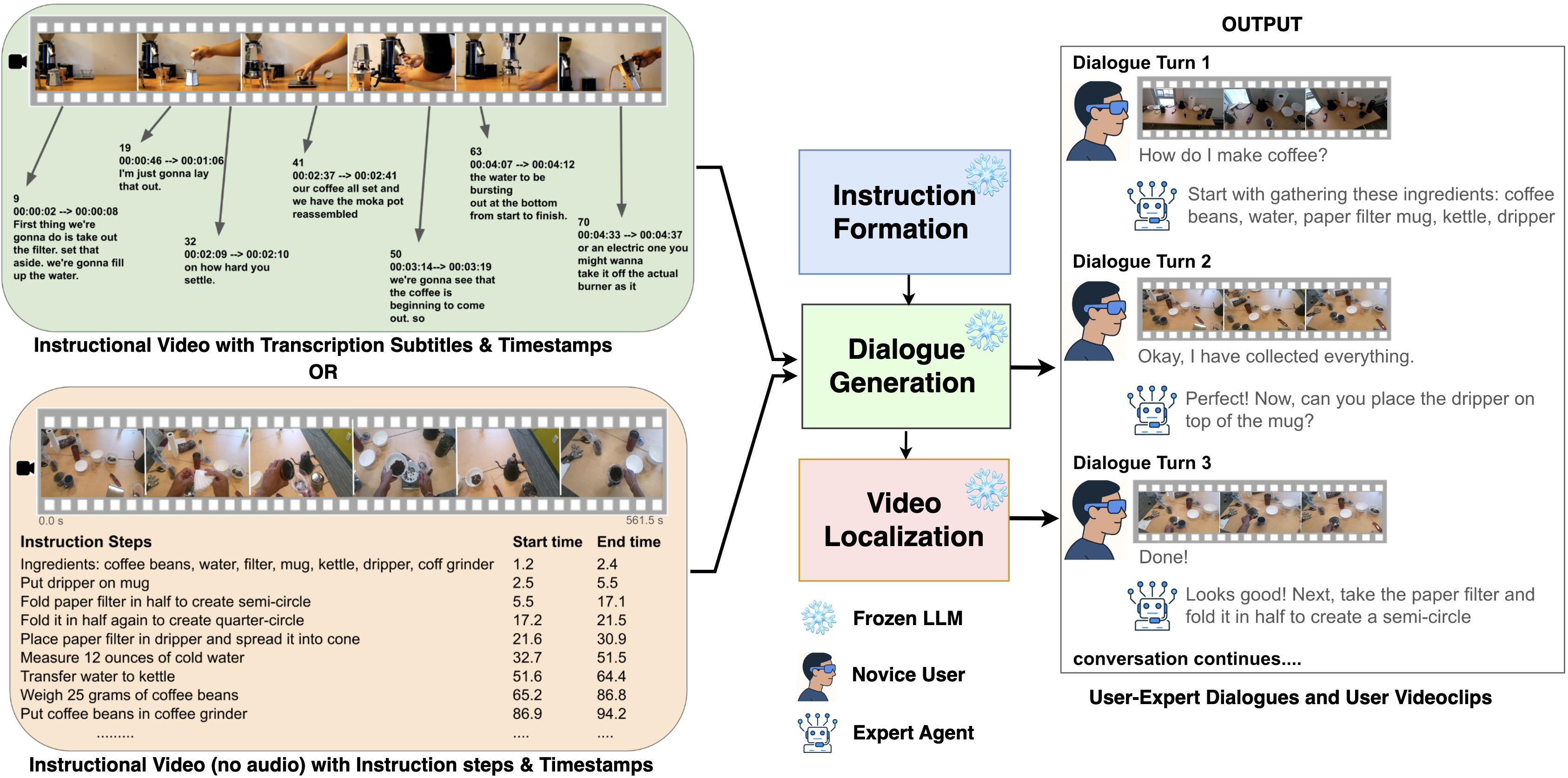}
         \caption{The input to our approach can be Top-left: an Instructional video with narration OR Bottom-left: an Instructional video (without narration) and step details. These are processed through the three step approach generating a conversation (with egocentric videos) between user and expert helper.}
         \label{fig:howtodiv_main_figure}
\end{figure*}
As illustrated in Fig.~\ref{fig:overview} and Fig.~\ref{fig:howtodiv_main_figure}, our pipeline comprises three stages: \emph{Instruction Formation}, \emph{Dialogue Generation}, and \emph{Video Localization}. We first briefly describe the underlying language model (Sec.~\ref{sec:gemma3}), formalize the problem setting (Sec.~\ref{sec:problem_formulation}), and then detail each stage of the pipeline (Secs.~\ref{sec:instruction_formation}-\ref{sec:video_localization}).

\subsection{Gemma 3}
\label{sec:gemma3}

Gemma 3~\cite{gemma3} is an open-weight multimodal model family supporting both text and image inputs via a SigLIP-based vision encoder. It offers a long context window (up to 128K tokens), enabling reasoning over lengthy transcripts, instruction lists, and multi-image inputs. Instruction-tuned variants provide strong multimodal and multilingual performance while remaining computationally lightweight, making them suitable for scalable dataset generation without finetuning.

\subsection{Problem Setting}
\label{sec:problem_formulation}
We consider an assistive scenario in which a user wears an augmented reality (AR) device equipped with sensors such as camera, microphone, and speaker. Through this interface, the user interacts with an AI agent that can perceive egocentric visual input and respond to spoken queries in real time. The agent’s objective is to guide the user through a procedural task by issuing step-wise, actionable instructions conditioned on the user’s progress.

At each turn, the agent receives a natural language query (e.g., 'What should I do next?') along with a short egocentric video clip capturing the user’s current state. The agent responds with the next instruction (e.g., 'Add two cups of water to the pan'). Constructing datasets to support this setting requires multi-turn dialogues in which each turn is grounded in both language and video.

We propose generating such data from egocentric instructional videos, which typically fall into one of the following categories:
\begin{enumerate}
    \item \textbf{Narrated instructional videos}, where an expert verbally explains each step while demonstrating the task. These videos are represented as a sequence of $N$ frames $V = (I_1, I_2, \dots, I_N)$ and subtitle segments $S = (S_1, S_2, \dots, S_M)$, where each $S_j = (s_j, t_{js}, t_{je})$ comprises of the narration text $s_j$, temporal boundaries $t_{js}$ and $t_{je}$.
    \item \textbf{Silent demonstration videos}, where an expert performs the task without narration, accompanied by step-level annotations $Q_j = (q_j, t_{js}, t_{je})$ describing the action $q_j$ and its temporal extent $t_{js}$ and $t_{je}$ .
\end{enumerate}

Our goal is to generate a conversation sequence $C = (c_1, c_2, \dots, c_P)$, where each turn $c_i = (u_i, e_i, v_i, t_{is}, t_{ie})$ consists of a user query $u_i$, an expert response $e_i$, video $v_i$ and the temporal span of the corresponding video segment. This representation aligns with the input-output structure expected by modern multimodal LLMs during training and inference.

\subsection{Instruction Formation}
\label{sec:instruction_formation}
The first stage extracts a complete, ordered set of task instructions, where each instruction corresponds to a single atomic action. Atomicity is enforced to ensure that each step can be executed and queried independently within a dialogue turn.

Due to heterogeneity in available supervision across datasets, we adopt dataset-specific extraction strategies. For narrated instructional videos (e.g., NIV~\cite{niv}), subtitle transcripts are obtained either from existing annotations or via automatic speech recognition. These transcripts are then processed by an LLM prompted to infer a concise, step-wise instruction list, augmenting implicit procedural knowledge when necessary.  

For datasets with fine-grained action annotations (e.g., EgoPER~\cite{egoper}), we apply post-processing to merge duplicate actions, normalize descriptions, and improve linguistic clarity. For other datasets not yet covered such as EGTEA~\cite{egtea}, semantically similar actions can be clustered, and non-essential actions (e.g., “open drawer”, “close lid”) filtered out to produce a minimal yet complete instruction set. The result is a canonical sequence of atomic instructions applicable across diverse task domains.

\subsection{Dialogue Generation}
\label{sec:dialogue_generation}
Given the ordered instruction list, we perform a second LLM inference to synthesize a multi-turn dialogue between a novice  and expert. The novice is assumed to have no prior access to the instruction list, while the expert has full task knowledge. Each atomic instruction is mapped to a single dialogue turn, consisting of a user query and a corresponding expert response.

Prompts are designed to generate both concise and longer conversational responses. To model realistic failure modes, we incorporate error annotations from datasets such as EgoPER~\cite{egoper}, which label erroneous steps and their corrective actions. Instruction steps corresponding to errors are marked with special tokens <user error>, enabling the LLM to generate expert responses that identify mistakes and provide corrective guidance. Additionally, procedural constraints, and nuances extracted from narrations are transformed into clarification questions, enriching the dialogue with domain-specific knowledge and natural interaction patterns.

\subsection{Video Localization}
\label{sec:video_localization}
The final stage temporally aligns each atomic instruction with a segment of the original instructional video, which is repurposed as the user’s egocentric visual input for that step. Localization strategies depend on the source dataset.  

For datasets with step-level temporal annotations (e.g., EgoPER, EGTEA), we can directly use the provided timestamps, aggregating segments when multiple actions are merged into a single instruction. For narrated instructional videos lacking explicit step boundaries, we rely on subtitle timestamps generated during transcription. Temporal boundaries are inferred during instruction extraction within the same LLM call. While more advanced temporal action localization methods could improve alignment, we leave their integration to future work. In practice, subtitle-based localization proved sufficient for turn-level grounding in our experiments.

\paragraph{Determinism and Stochasticity.}


While the pipeline relies on LLM inference, its structure is governed by fixed constraints. Instruction formation deterministically produces an ordered, atomic step sequence given the same transcripts or annotations. Dialogue generation enforces a one-to-one mapping between steps and turns, with stochasticity limited to surface-level linguistic variation. Video localization is deterministic when temporal annotations are available, and otherwise uses inferred subtitle timestamps. As a result, identical prompts yield consistent procedural structure and video alignment, with variation confined to lexical expression.

Overall, the MDC pipeline performs instruction canonicalization, turn-structured dialogue synthesis, and temporal grounding, yielding a reproducible mapping from instructional videos to multimodal conversational data.

\section{Experiments}
\label{sec:experiment}

\subsection{Implementation Details}
We use Gemma-3~\cite{gemma3} as the backbone language model for instruction extraction, dialogue synthesis, and video-step labeling due to its strong instruction-following and long-context capabilities. Unless stated otherwise, we employ the instruction-tuned 27B variant during dataset generation. All experiments are conducted using Fully Sharded Data Parallel (FSDP) training on TPU v3 hardware with 8 compute nodes and 4 Dragonfish chips.

We partition the dataset into training, validation, and test splits using a 70\% / 10\% / 20\% ratio, stratified by task type and user category. This results in 355 training sessions, 44 validation sessions, and 108 test sessions. All quantitative results are reported on the held-out test set.
\paragraph*{Dialogue Diversity.}
To encourage diversity in generated interactions, we adopt random sampling with a temperature of 1.5, nucleus sampling with top-$p=0.9$. This choice yields varied user utterances and expert response styles, as illustrated in Fig.~\ref{fig:howtodiv_main_figure}. We produce a single response per dialogue turn using autoregressive decoding without reranking. For EgoPER sessions, we generate two variants of each dialogue corresponding to regular (usually verbose) and concise user speech, increasing linguistic coverage.

\paragraph*{Inference Setup.}
For expert response generation at inference time, we benchmark Gemma-3-4B ~\cite{gemma3} and Qwen2.5-VL-7B~\cite{qwen2.5} and evaluate two prompt configurations:
\begin{itemize}
    \item \textbf{History-only}: the model receives the prior conversation and must infer subsequent steps using its internal knowledge.
    \item \textbf{History + Steps}: the full instruction list is appended to the system prompt, enabling explicit reasoning over task structure.
\end{itemize}
The latter serves as an upper-bound reference, while the former reflects a more realistic deployment scenario.
\setlength{\textfloatsep}{8pt}
\setlength{\intextsep}{2pt}
\begin{figure}[ht]
     \centering   
     \includegraphics[width=0.55\textwidth]
     {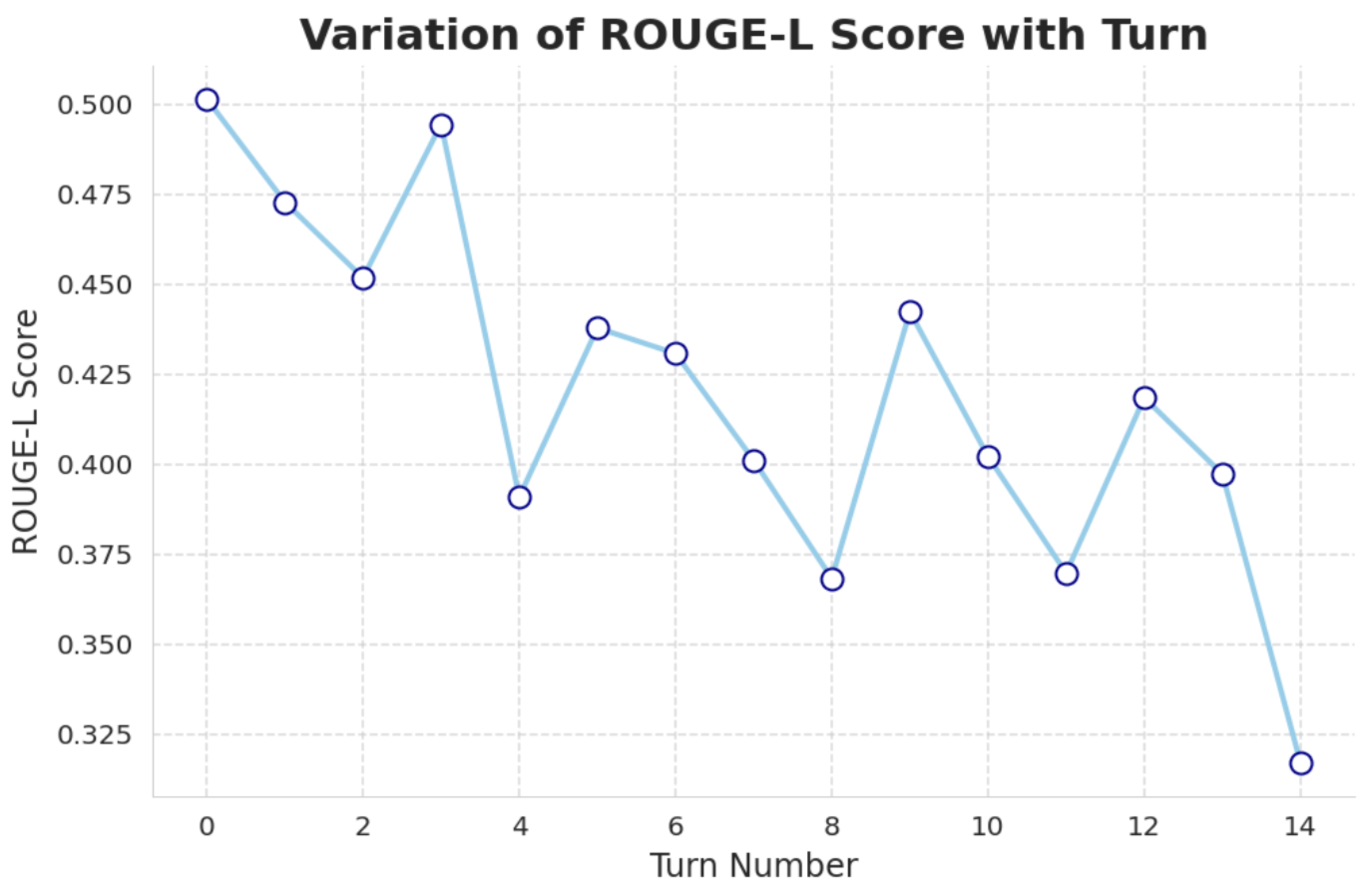}
     \caption{Variation of mean ROUGE-L score with increasing dialogue turns for HowToDIV sessions demonstrating compounding error in long-horizon procedural dialogue}
     \label{fig:bleu_turnwise}
\end{figure}

\subsection{Evaluation Metrics}
We evaluate expert response generation using three complementary metrics.


\textbf{LLM-as-a-Judge.}
Following prior work~\cite{llmasjudge,mllmasjudge}, we use an LLM-based evaluation protocol to assess semantic response quality. A Gemma-3 12B model is prompted with the task name, user query, reference response, and generated output; it produces a brief justification along with a scalar score on a 1-5 scale based on correctness, relevance, completeness, and helpfulness. To mitigate ordering bias, reference and generated responses are presented in randomized order. Scores are averaged across all test turns. While LLM-based evaluation may inherit model biases, it enables scalable semantic assessment of open-ended procedural guidance, which is difficult to capture using purely lexical metrics.

\textbf{ROUGE.}
ROUGE~\cite{rouge} measures recall-oriented lexical overlap between generated and reference responses. We report ROUGE-1, ROUGE-2, and ROUGE-L, capturing unigram recall, bigram recall, and longest common subsequence overlap, respectively.

\textbf{BLEU.}
BLEU~\cite{bleu} measures n-gram precision between generated and reference responses using a brevity-penalized geometric mean, with scores normalized to $[0,1]$.

For reference-based metrics (BLEU and ROUGE), scores are computed at the dialogue-turn level and averaged over the test set. Since LLM-as-a-Judge captures high-level semantic adequacy, ROUGE emphasizes content coverage, and BLEU reflects local precision and fluency, we report all three metrics to capture complementary aspects of expert response quality.

\begin{figure}[ht]
     \centering   
    \includegraphics[width=0.7\textwidth]
     {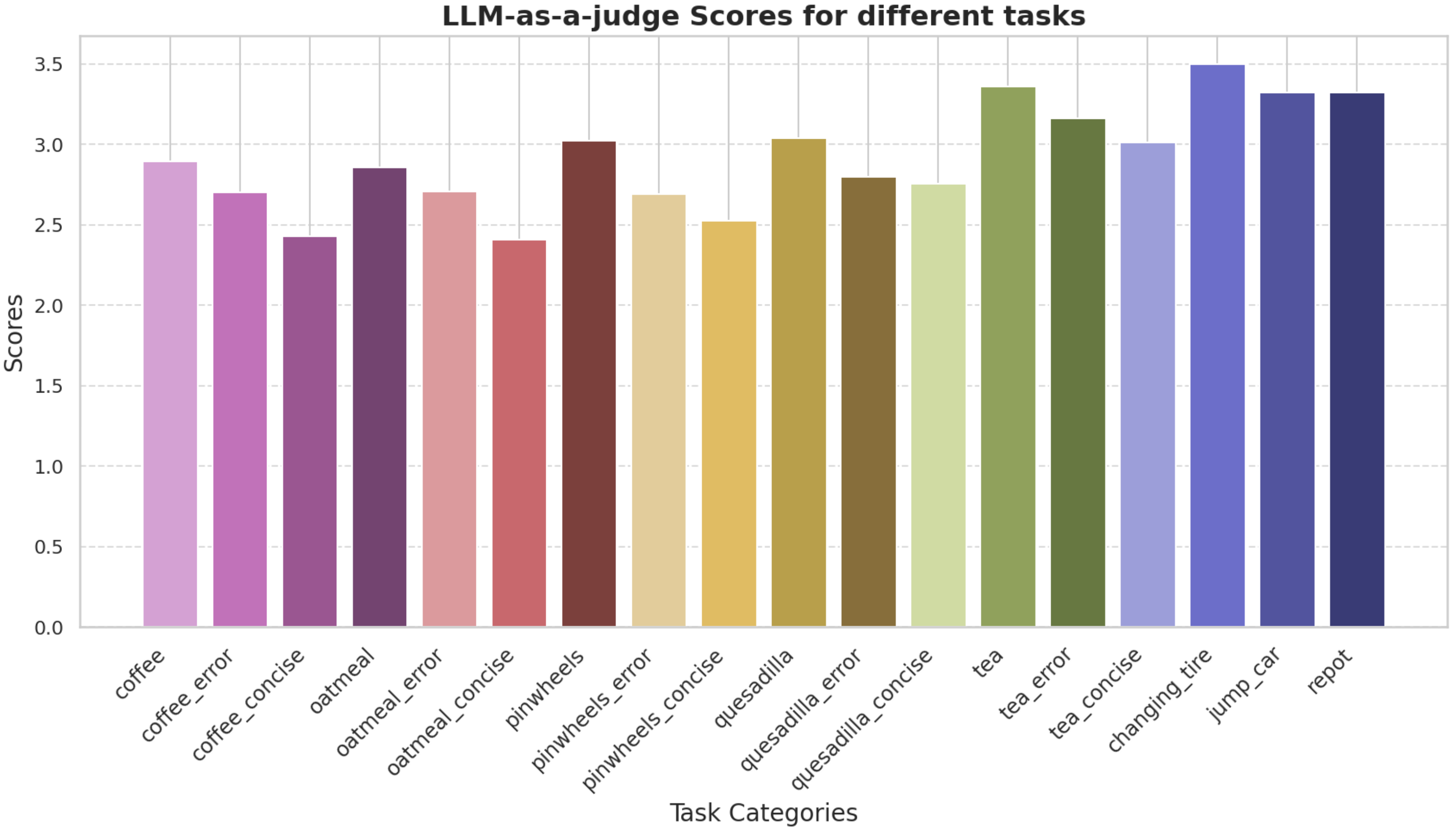}
     \caption{Variation of LLM-as-a-Judge scores across different tasks, speech-styles and action-categories for Gemma-3-4B expert responses.}
     \label{fig:llmj_taskwise}
\end{figure}
\setlength{\textfloatsep}{6pt}
\setlength{\intextsep}{6pt}
\begin{figure}[ht]
     \centering   
     \includegraphics[width=0.5\textwidth]
     {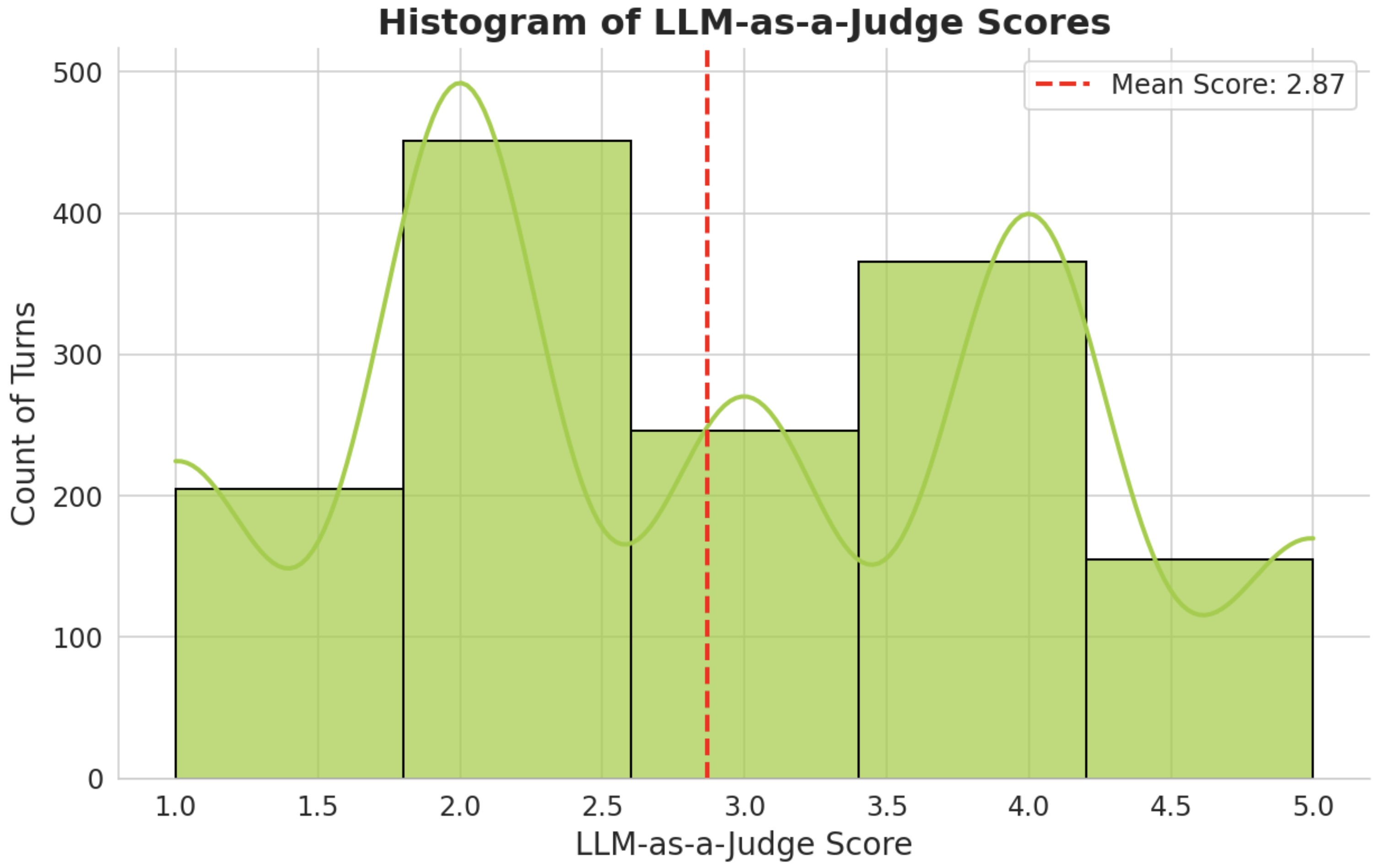}
     \caption{Histogram of LLM-as-a-Judge scores on HowToDIV for History-only prompt on Gemma-3-4B responses.}
     \label{fig:llm_aaj_scores}
\end{figure}

\subsection{Results and Observations}
\label{sec:main_results}

\begin{table}[t]
\setlength{\abovecaptionskip}{6pt}
\centering
\footnotesize
\setlength{\tabcolsep}{6pt}
\renewcommand{\arraystretch}{1.15}

\begin{tabular}{lccccc}
\toprule
\textbf{Method} & \textbf{BLEU} & \textbf{LLM-Judge} & \textbf{ROUGE-1} & \textbf{ROUGE-2} & \textbf{ROUGE-L} \\
\midrule
Gemma-3-4B (History-only) 
& 0.321 & 2.870 & 0.325 & 0.125 & 0.270 \\

Gemma-3-4B (History + Steps) 
& \textbf{0.457} & 4.101 & \textbf{0.489} & \textbf{0.268} & \textbf{0.429} \\

Qwen2.5-VL-7B (History-only) 
& 0.281 & 2.995 & 0.310 & 0.105 & 0.220 \\

Qwen2.5-VL-7B (History + Steps) 
& 0.441 & \textbf{4.232} & 0.466 & 0.199 & 0.356 \\

\bottomrule
\end{tabular}

\caption{Evaluation on HowToDIV comparing Gemma-3 and Qwen2.5 configurations. The 12B Gemma-3 variant is used for the LLM-as-a-Judge metric.}
\label{tab:howtodiv_metrics}
\end{table}

\begin{table}[t]
\setlength{\abovecaptionskip}{6pt}
\centering
\footnotesize
\setlength{\tabcolsep}{6pt}
\renewcommand{\arraystretch}{1.15}

\begin{tabular}{lccccc}
\toprule
\textbf{User Category} & \textbf{BLEU} & \textbf{LLM-Judge} & \textbf{ROUGE-1} & \textbf{ROUGE-2} & \textbf{ROUGE-L} \\
\midrule
Concise-Follow  & 0.307 & 2.616 & 0.309 & 0.116 & 0.258 \\
Regular-Follow  & \textbf{0.335} & \textbf{3.066} & \textbf{0.342} & \textbf{0.135} & \textbf{0.282} \\
Regular-Error   & 0.304 & 2.816 & 0.304 & 0.109 & 0.257 \\
\bottomrule
\end{tabular}

\caption{Performance across user speech styles and action types on HowToDIV using Gemma-3 with the History-only prompt configuration.}
\label{tab:usertype_metrics}
\end{table}

\begin{figure}[ht]
     \centering   
     \includegraphics[width=0.65\textwidth]
     {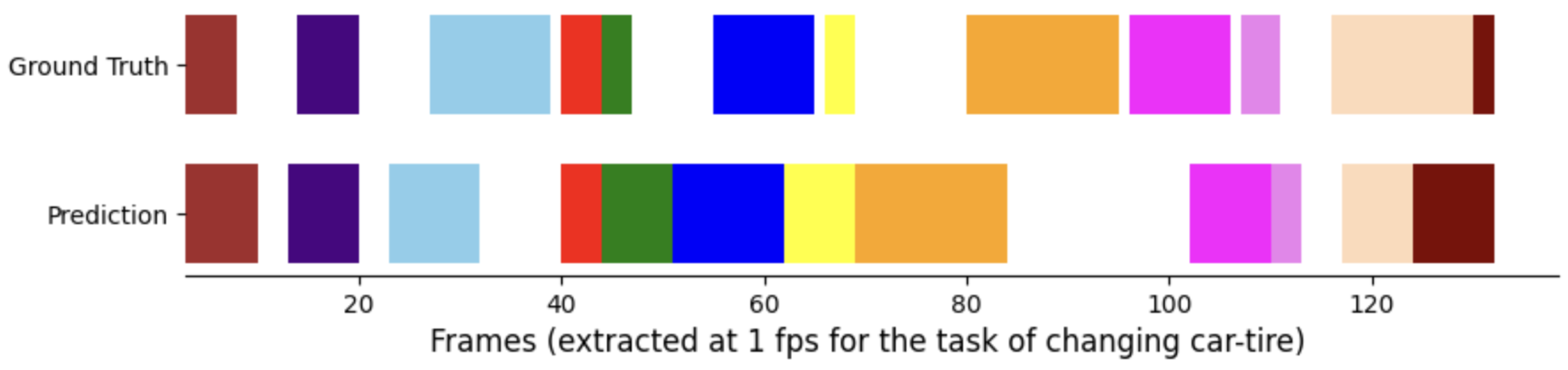}
     \caption{Temporal action localization for changing car tire; each color denotes a step}
     \label{fig:action_segmentation_car}
\end{figure}


Tab.~\ref{tab:howtodiv_metrics} compares the different prompting strategies and models. Across all metrics, History + Steps configuration substantially outperforms the History-only setup for both Gemma-3 and Qwen-2.5, highlighting the importance of explicit task structure. Without access to step-level instructions, the model relies solely on prior knowledge, which frequently leads to hallucinations and domain-specific errors.  
Gemma-3-4B performs better on BLEU and ROUGE metrics, whereas Qwen2.5-VL-7B yields a marginally higher LLM-as-a-Judge score.

Figure~\ref{fig:llm_aaj_scores} shows that the median LLM-as-a-Judge score for History-only setup with Gemma-3 is 3.0. Shown in Fig.~\ref{fig:bleu_turnwise}, performance consistently degrades as dialogue length increases, indicating compounding error and reduced long-horizon reasoning. 

Tab.~\ref{tab:usertype_metrics} and Fig.~\ref{fig:llmj_taskwise} analyze performance across user speech styles and action types. Regular-speech interactions consistently outperform concise-speech variants across metrics. Moreover, dialogues in which users follow instructions yield significantly higher scores than those involving user errors, reflecting the increased complexity of corrective reasoning. This trend holds across most task categories. Finally, we also demonstrate the quality of temporal step localization in Figure~\ref{fig:action_segmentation_car} comparing predicted action segments against human annotations for a representative NIV recording.

\section{Conclusion}
\label{sec:conclusion}
We introduced a prompt-based framework for converting egocentric monologue instructional videos into multi-turn, two-person conversations for procedural task assistance. By leveraging the reasoning and language generation capabilities of large language models, our approach synthesizes natural interactions grounded in temporally localized egocentric videos, providing a scalable alternative to costly in-field data collection and manual annotation. Using this framework, we presented HowToDIV, a
multimodal dataset of expert–novice procedural dialogues grounded in egocentric video. It addresses a critical gap in existing resources for task-oriented, multimodal conversations and enables benchmarking of instructional dialogue agents. We establish baseline results using Gemma-3 and Qwen-2.5 models across multiple evaluation metrics, providing a reference point for future work. We rely on subtitle-based localization for turn-level grounding and leave learned action localization to future work. We believe this dataset will facilitate progress toward interactive AI agents capable of assisting users through complex procedures in real-time settings.

\FloatBarrier


\begin{thebibliography}{8}
\bibitem{charades}
Sigurdsson, G.A., Varol, G., Wang, X., Farhadi, A., Laptev, I., Gupta, A.:
Hollywood in homes: Crowdsourcing data collection for activity understanding.
In: European Conference on Computer Vision (ECCV),
pp. 510--526. Springer, Cham (2016)

\bibitem{alfred}
Shridhar, M., Thomason, J., Gordon, D., Bisk, Y., Han, W., Mottaghi, R.,
Zettlemoyer, L., Fox, D.:
Alfred: A benchmark for interpreting grounded instructions for everyday tasks.
In: Proceedings of the IEEE/CVF Conference on Computer Vision and Pattern Recognition (CVPR),
pp. 10740--10749 (2020)

\bibitem{holoassist}
Wang, X., Kwon, T., Rad, M., Pan, B., Chakraborty, I., Andrist, S., Bohus, D.,
Feniello, A., Tekin, B., Frujeri, F.V., Joshi, N., Pollefeys, M.:
HoloAssist: An egocentric human interaction dataset for interactive AI assistants
in the real world.
In: Proceedings of the IEEE/CVF International Conference on Computer Vision (ICCV),
pp. 20270--20281 (2023)







\bibitem{epictent}
Jang, Y., Sullivan, B., Ludwig, C., Gilchrist, I., Damen, D., Mayol-Cuevas, W.:
Epic-Tent: An egocentric video dataset for camping tent assembly.
In: Proceedings of the IEEE/CVF International Conference on Computer Vision Workshops (ICCVW)
(2019)

\bibitem{egtea}
Li, Y., Liu, M., Rehg, J.M.:
In the eye of the beholder: Joint learning of gaze and actions in first person video.
In: European Conference on Computer Vision (ECCV),
pp. 619--635. Springer, Cham (2018)

\bibitem{bleu}
Papineni, K., Roukos, S., Ward, T., Zhu, W.-J.:
BLEU: A method for automatic evaluation of machine translation.
In: Proceedings of the 40th Annual Meeting of the Association for Computational Linguistics (ACL),
pp. 311--318 (2002)


\bibitem{rouge}
Lin, C.-Y.:
ROUGE: A package for automatic evaluation of summaries.
In: Text Summarization Branches Out,
pp. 74--81 (2004)

\bibitem{llmasjudge}
Zheng, L., Chiang, W.-L., Sheng, Y., Zhuang, S., Wu, Z., Zhuang, Y., Lin, Z.,
Li, Z., Li, D., Xing, E.:
Judging LLM-as-a-judge with MT-Bench and Chatbot Arena.
Advances in Neural Information Processing Systems \textbf{36},
46595--46623 (2023)

\bibitem{mllmasjudge}
Chen, D., Chen, R., Zhang, S., Wang, Y., Liu, Y., Zhou, H., Zhang, Q.,
Wan, Y., Zhou, P., Sun, L.:
MLLM-as-a-judge: Assessing multimodal LLM-as-a-judge with vision--language benchmarks.
In: Proceedings of the 41st International Conference on Machine Learning (ICML) (2024)

\bibitem{niv}
Alayrac, J.-B., Bojanowski, P., Agrawal, N., Sivic, J., Laptev, I., Lacoste-Julien, S.:
Unsupervised learning from narrated instruction videos.
In: Proceedings of the IEEE Conference on Computer Vision and Pattern Recognition (CVPR),
pp. 4575--4583 (2016)

\bibitem{egoper}
Lee, S.-P., Lu, Z., Zhang, Z., Hoai, M., Elhamifar, E.:
Error detection in egocentric procedural task videos.
In: Proceedings of the IEEE/CVF Conference on Computer Vision and Pattern Recognition (CVPR),
pp. 18655--18666 (2024)

\bibitem{wikihow}
Koupaee, M., Wang, W.Y.:
WikiHow: A large-scale text summarization dataset.
arXiv:1810.09305 (2018)

\bibitem{youcook2}
Zhou, L., Louis, N., Corso, J.J.:
Weakly supervised video object grounding from text by loss weighting and object interaction.
arXiv:1805.02834 (2018)

\bibitem{gemma3}
Gemma Team, Kamath, A., Ferret, J., Pathak, S., Vieillard, N., Merhej, R.,
Perrin, S., Matejovicova, T., Ram{\'e}, A., Rivi{\`e}re, M.:
Gemma 3 technical report.
arXiv:2503.19786 (2025)

\bibitem{qwen2.5}
Qwen Team:
Qwen2.5: A party of foundation models.
\url{https://qwenlm.github.io/blog/qwen2.5/} (2024)


\bibitem{assembly101}
Sener, F., Chatterjee, D., Shelepov, D., He, K., Singhania, D., Wang, R., Yao, A.:
Assembly101: A large-scale multi-view video dataset for understanding procedural activities.
In: Proceedings of the IEEE/CVF Conference on Computer Vision and Pattern Recognition (CVPR) (2022)

\bibitem{crosstask}
Zhukov, D., Alayrac, J.-B., Cinbis, R.G., Fouhey, D., Laptev, I., Sivic, J.:
Cross-task weakly supervised learning from instructional videos.
In: Proceedings of the IEEE/CVF Conference on Computer Vision and Pattern Recognition (CVPR),
pp. 3537--3545 (2019)

\bibitem{epickitchen}
Damen, D., Doughty, H., Farinella, G.M., Fidler, S., Furnari, A., Kazakos, E.,
Moltisanti, D., Munro, J., Perrett, T., Price, W., Wray, M.:
Scaling egocentric vision: The EPIC-KITCHENS dataset.
In: European Conference on Computer Vision (ECCV). Springer, Cham (2018)

\bibitem{coin}
Tang, Y., Ding, D., Rao, Y., Zheng, Y., Zhang, D., Zhao, L., Lu, J., Zhou, J.:
COIN: A large-scale dataset for comprehensive instructional video analysis.
In: Proceedings of the IEEE/CVF Conference on Computer Vision and Pattern Recognition (CVPR),
pp. 1207--1216 (2019)

\bibitem{howto100m}
Miech, A., Zhukov, D., Alayrac, J.-B., Tapaswi, M., Laptev, I., Sivic, J.:
HowTo100M: Learning a text--video embedding by watching hundred million narrated video clips.
In: Proceedings of the IEEE/CVF International Conference on Computer Vision (ICCV),
pp. 2630--2640 (2019)

\bibitem{msrvtt}
Xu, J., Mei, T., Yao, T., Rui, Y.:
MSR-VTT: A large video description dataset for bridging video and language.
In: Proceedings of the IEEE Conference on Computer Vision and Pattern Recognition (CVPR),
pp. 5288--5296 (2016)

\bibitem{ego4d}
Grauman, K., Westbury, A., Byrne, E., Chavis, Z., Furnari, A., Girdhar, R.,
Hamburger, J., Jiang, H., Liu, M., Liu, X.:
Ego4D: Around the world in 3,000 hours of egocentric video.
In: Proceedings of the IEEE Conference on Computer Vision and Pattern Recognition (CVPR),
pp. 18995--19012 (2022)






\end{thebibliography}
\end{document}